\documentclass{article}

\usepackage[square,numbers]{natbib}
\usepackage[preprint]{neurips_2024}

\usepackage[utf8]{inputenc} 
\usepackage[T1]{fontenc}    
\usepackage{hyperref}       
\usepackage{url}            
\usepackage{booktabs}       
\usepackage{amsfonts}       
\usepackage{nicefrac}       
\usepackage{microtype}      
\usepackage{xcolor}         
\usepackage[normalem]{ulem}

\usepackage{paralist}
\usepackage{subfloat}
\usepackage{subfig}
\usepackage{graphicx}
\usepackage{tabularx}
\usepackage{array}
\usepackage{multirow}
\usepackage{multicol}
\usepackage{amsmath}
\usepackage{booktabs}
\usepackage{bbm}
\usepackage{algorithm}
\usepackage{algpseudocode}
\usepackage{xspace}
\newcommand{\model}{\textsc{SURP}\xspace}
\usepackage[textsize=tiny]{todonotes}

\newcommand{\allnotes}[1]{}
\newcommand{\ignore}[1]{}

\renewcommand{\allnotes}[1]{\textit{#1}}

\title{Adaptive Pre-training Data Detection for Large Language Models via Surprising Tokens}

\author{%
  Anqi Zhang \\
  \And
  Chaofeng Wu \\
}

\begin{document}

\maketitle

\begin{abstract}
While large language models (LLMs) are extensively used, there are raising concerns regarding privacy, security, and copyright due to their opaque training data, which brings the problem of detecting pre-training data on the table.
Current solutions to this problem leverage techniques explored in machine learning privacy such as Membership Inference Attacks (MIAs), which heavily depend on LLMs' capability of verbatim memorization. 
However, this reliance presents challenges, especially given 
the vast amount of training data and the restricted number of effective training epochs. 
In this paper, we propose an adaptive pre-training data detection method which alleviates this reliance and effectively amplify the identification. 
Our method adaptively locates \textit{surprising tokens} of the input. 
A token is surprising to a LLM if the prediction on the token is "certain but wrong", which refers to low Shannon entropy of the probability distribution and low probability of the ground truth token at the same time.
By using the prediction probability of surprising tokens to measure \textit{surprising}, the detection method is achieved based on the simple hypothesis that seeing seen data is less surprising for the model compared with seeing unseen data. 
The method can be applied without 
any access to the the pre-training data corpus or additional training like reference models.
Our approach exhibits a consistent enhancement compared to existing methods in diverse experiments conducted on various benchmarks and models, achieving a maximum improvement of 29.5\%. We also introduce a new benchmark Dolma-Book developed upon a novel framework, which employs book data collected both before and after model training to provide further evaluation. 

\end{abstract}

\section{Introduction}
\label{sec:intro}
Large language models(LLMs) are experiencing immense success, with their remarkable performance and widespread popularity driving their research and deployment across a multitude of applications~\cite{li2024pre, li2024flexkbqa, dong2023towards, achiam2023gpt, touvron2023llama, huang2024free, thoppilan2022lamda}. Alongside this, concerns regarding the potential negative impacts on privacy~\cite{pan2020privacy, Brown2022WhatDI, li2023multi,yu2021differentially}, security~\cite{yao2024survey, mozes2023use, carlini2021extracting, qi2024visual} and copyright~\cite{chu2024protect, kirchenbauer2023watermark, vyas2023provable} of LLMs are also gathering widespread attention. 
Lacking of transparency regarding training data~\cite{shi2024detecting}, as one significant challenge, leads to difficulty in model evaluation~\cite{oren2024proving, magar-schwartz-2022-data}, and a series of problems including potential leaks of personal privacy data~\cite{kim2024propile, 10179300} and infringements upon copyrighted data~\cite{chang2023speak,henderson2023foundation}.

In tackling this challenge, a group of works is delving into the problem of pre-training data detection, which aims to determine whether the target LLM was pre-trained on a given input without prior knowledge of the pre-training data and only with black-box access to it. This is in line with Membership Inference Attacks (MIAs)~\cite{hu2022membership}, a problem extensively studied in machine learning privacy and gaining attention under LLMs domain.
MIAs exploit the model's tendency to overfit the training data, resulting in low loss values for that data. Typically, prevalent strategies for pre-training data detection based on MIAs heavily rely on the \textit{verbatim memorization}~\cite{hartmann2023sok} capabilities of LLMs. However, this reliance poses significant challenges~\cite{duan2024membership}, especially given the current scenario where pre-trained LLMs generally have dramatically large scale training sets and are only trained on the data for around one epoch.

\begin{figure*}[t]
    \centering
    \includegraphics[width=\textwidth]{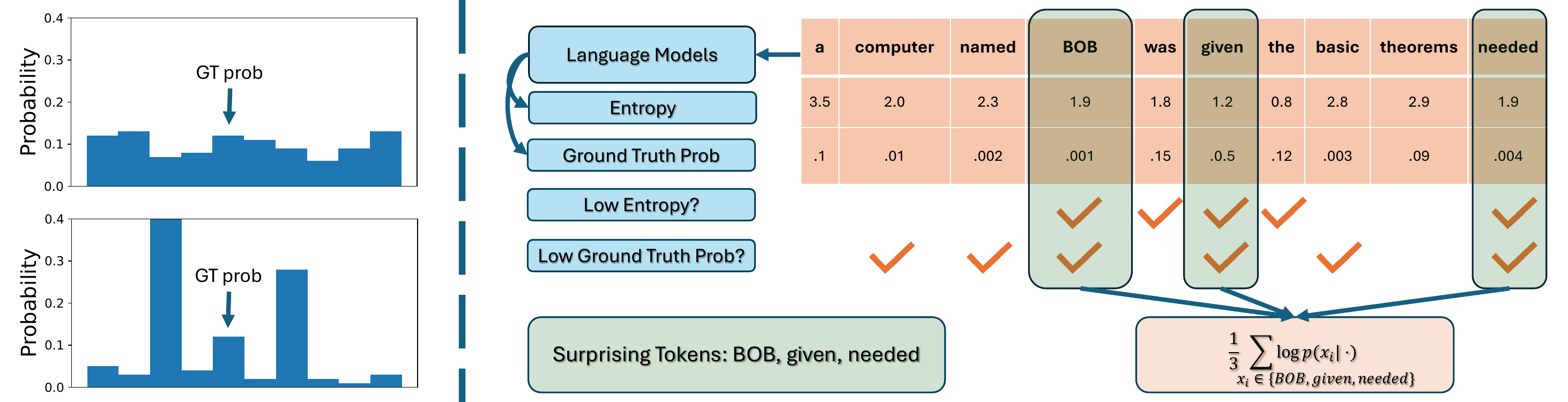}
    \caption{
    Overview of \model. 
    The left part illustrates surprising token (bottom left) and unsurprising token (upper left). The x-axis represents words in vocabulary (token candidates), the y-axis represents the probability that the model assigns to each word. A \textit{surprising token} should satisfy both (a) the probability distribution is not flat (low entropy) and (b) the probability of ground truth token (GT prob in figure) is relatively low. 
    The right part shows the flow of \model. Given an input, we can get the entropy and ground truth probability at each index. Then we use the average ground truth probability of surprising tokens as the score of the input, to determine whether it is seen or not.}
    \label{fig:overview}
\vspace{-15pt}
\end{figure*}

In this work, we aim to exploring the issue from a different perspective: if the verbatim memorization of arbitrary pre-training data is unreliable for the model, is it possible for people to automatically locate representative places
of the input, which effectively amplify the distinctions between the model's \textit{seen} data (i.e., pre-training data) and \textit{unseen} data (i.e., non-training data)? 
Distinguished from prior works, our method wants to fill in this gap and to conduct detection through method that alleviates the reliance on the plain input memory as much as possible. 

We propose an adaptive pre-training data detection method \model, which locate \textit{surprising tokens} of a given input and achieve detection based on the hypothesis that 
seeing seen data is less \textit{surprising} for model when compared with seeing unseen data.
For the input, we first adaptively locate \textit{surprising tokens}. The surprising tokens are inspired by Shannon Entropy\cite{shannon1948mathematical} in information theory, empirical observations(Figure \ref{fig:demo}), and the nature of surprise. 
Surprising happens when people are sure about the answer of a question, but the revealed true answer is different from the answer in mind. 
We transfer this definition of surprise to LLMs, and define \textit{surprising token} as tokens in the input where the LLM is sure about what the prediction should be, but the probability that the LLM assigned to the ground truth token is low.
We use Shannon entropy to measure how sure the LLMs think of what the next token is, as it measures the uncertainty of a distribution. 
Lower entropy means the probability distribution is more concentrated, and there’s less uncertainty about the outcome. 
Thus, \textit{surprising tokens} refers to low Shannon entropy of the probability distribution and low probability of the ground truth token at the same time.
Further, to measure how \textit{surprising} the LLM is, we use the average prediction probability of surprising tokens: 
the lower the average probability, the more surprising the LLM. Based on the hypothesis, LLM will have higher average probability on seen data compared with on unseen data. 
To summarize the process, given an input, our method first adaptively locates the surprising token of input, then calculate the average probability that the LLM assigned to the surprising tokens, finally detect whether the input is seen or not based on the average probability. Our approach can be applied without any prior knowledge about the pre-training data corpus or additional training like reference models.

We construct a new benchmark Dolma-Book, to enlarge the limited number of datasets for pre-training data detection problem. Dolma-Book is developed upon a novel LLM framework – Open Language Model (OLMo)~\cite{OLMo}. We employs book data from Project Gutenberg collected before and after model training(i.e., based on a cutoff date) to provide further evaluation of the detection methods. Since the considerable length of the raw book data, we collect and category three datasets from the \textit{head}, \textit{middle} and \textit{tail} segmentations of the book to serve for a better evaluation of detection performance across various text positions within the long-context book data. Procedure of data collection and categorization is automatic and convenient to extend for future studies. 
We empirically evaluate the performance of our approach across various language models and data benchmarks. 
Our approach consistently outperforms existing methods across three benchmarks and different families of LLMs. Further analysis demonstrate our method maintains advantages on deduplicated model and inputs of different length variations. 
Code and benchmark are available here: make public after acceptance.

We summarize our contributions below:
\begin{compactitem}
    \item We propose to identify surprising tokens (\S~\ref{subsec:surpring-tokens}) instead of blindly believing in the model's memory of each individual token, due to the challenges and unreliability of verbatim memorization (\S~\ref{sec:related-works}).
    \item We construct an improved pre-training data detection approach (\S~\ref{subsec:detection-method}) through adaptively surprising tokens identification and designed score computation.
    \item We construct a book data benchmark for further pre-training detection problem (\S~\ref{subsec:eval-setups}) and provide extensive empirical demonstration of black-box detections for various LLMs and benchmarks (\S~\ref{sec:eval}). 
\end{compactitem}

\vspace{-.05in}
\section{Related Work}
\label{sec:related-works}
\vspace{-.05in}
We discuss related works on pre-training data detection for large language models. The problem of detecting pre-training data aims to determine whether a language model was trained on the given text input when black-box access to the LLM without knowing the pre-training data. Its necessity has received increasing attention from various aspects. 

In the field of traditional machine learning privacy, there has been well explored research on membership inference~\cite{shokri2017membership, hu2022membership, yeom2018privacy, mireshghallah2022quantifying, carlini2022membership, choquette2021label}. Since the current problem is in line with it, some prior works use MIA technologies to solve detection problem in LLM domain. A simple and frequently employed method is the LOSS attack~\cite{yeom2018privacy}, which utilizes the model's computed loss and classifies the inputs as part of the training set if their loss values fall below a specific threshold. 
Some work~\cite{carlini2022membership, watson2022on} claims this kind of attacks have high false positive rate (i.e., non-member samples are often erroneously predicted as members), 
and \textit{difficulty calibration}~\cite{watson2022on} is proposed to overcome this. 
Since then, other reference-based methods~\cite{carlini2021extracting, mireshghallah-etal-2022-quantifying} are proposed to consider the intrinsic complexity of the target data by calibrating the loss of target model with another reference model. These methods have to train other models on the same dataset or data from same distribution. Instead of training reference models, \cite{mattern-etal-2023-membership} craft textual \textit{neighboring samples} through data augmentation and leverages an estimation of the loss curvature to detect membership. More recently, \cite{shi2024detecting} selects tokens which have the lowest likelihood to compute the score for detection.

Applying MIAs in detecting pre-training data of LLMs has its own challenges.
Since traditional MIAs exploit the model's overfitting on the training data (i.e., model tends to obtain low loss values on training data), basically, the commonly-used methods 
heavily rely on LLMs' capabilities of \textit{verbatim memorization}~\cite{carlini2023quantifying,ippolito2023preventing, hartmann2023sok, NEURIPS2022_fa0509f4}. 
However, existing work~\cite{duan2024membership} claims challenges in solving data detection problem simply based on memorization of input, 
as the standard practice of training LLMs is training model for one epoch with massive training corpus, which makes overfitting impractical.
Not relying on memorizing each word input but uses some narrow concepts to screen, \cite{chang2023speak} proposes \textit{name cloze} for data archaeology to infer books that are known to ChatGPT/GPT-4. 
Although insightful design, the scenario is very limited. Usually, inferring membership based on named-entity is not robust due to the unsure information rarity. 

In parallel to pre-training data detection, there are some related but not completely overlapping issues. \cite{oren2024proving, golchin2024time, magar-schwartz-2022-data, Blevins2022LanguageCH, Dodge2021DocumentingLW} study the data contamination problem, where the pre-training dataset contains various evaluation benchmarks. \cite{274574, henderson2023foundation, pmlr-v202-yu23g} work on the data extraction problem, which aim to recover individual training examples through prompting attacks, especially the personally identifiable information(PII)~\cite{10179300}. 
These methods have limited use in pre-training data detection problem, as they require the specific dataset structure or data types.

\vspace{-.05in}
\section{Preliminary}
\label{sec:prelim}
\vspace{-.05in}
We first define the terminology we use, formalize the problem, and state our assumptions about the knowledge and capabilities. 
We follow the standard definition of the existing pre-training data detection in LLMs:

\vspace{-.1in}
\paragraph{Autoregressive language models.} We focus on the autoregressive language models $\mathcal{M}$, which are trained to predict the next text token based on the input sequence of previous tokens (i.e., different from bidirectional masked language models such as BERT~\cite{devlin-etal-2019-bert}). We use $p(\cdot | x_1, x_2, ..., x_{i-1})$ to represent the probability distribution of an model $\mathcal{M}$ on generating next token when given an input sequence ${x_1, x_2, ..., x_{i-1}}$. This probability distribution is among a token vocabulary, denoted as $\mathcal{V}$, and for each $v_j \in \mathcal{V}$, denote $p(v_j\mid x_1, .., x_{i-1})$ as the probability of next token to be $v_j$. 
\vspace{-.1in}
\paragraph{Clarify names and definitions.} We want to clarify the following names used in paper for better understanding. Given an input sequence $x_1, x_2, \dots, x_N$, each of $x_i$ is called \textit{token} (i.e., ground-truth token); \textit{index} (i.e, $1,2,..N$) is used to state the location of $x_i$. 
We call the possible generated tokens (i.e., $\forall~v_j \in \mathcal{V}$) at a specific index as \textit{token candidates}.
\vspace{-.1in}
\paragraph{Problem definition.} Given a data point $\mathbf{x} = x_1, x_2,..,x_N$ and a pre-trained language model $\mathcal{M}$, the goal of pre-training data detection problem is to infer whether $\mathbf{x}$ belongs to the training dataset $\mathcal{D}$ for this model $\mathcal{M}$. 
The detection process $\mathcal{F}$ is achieved by computing a score $f(\mathbf{x}, \mathcal{M})$ and decided through a case-dependent threshold $\lambda$. Formally, $\mathbf{x}: \mathcal{F}(f(\mathbf{x}, \mathcal{M}); \lambda) \mapsto \{0, 1\}$. 
Following previous works, people care about the threshold-independent metric, area under the ROC curve, for the aggregate measure of performance. 
\vspace{-.1in}
\paragraph{Knowledge and capabilities.} We assume the detector can access to the output statistics of the model $\mathcal{M}$, i.e., probabilities, but no access to model parameters, gradients as well as the pre-training data $\mathcal{D}$. Also, we assume no access to the pre-training data distribution $\mathbbm{D}$ for any training of calibration model. 

\section{Adaptive Pre-training Data Detection through Surprising Tokens}
In this section, we provide the motivation and detailed description of our adaptive pre-training data detection method \model, that using token probabilities at representative index for detection. 

Since current state-of-art pre-trained LLMs are trained with billions and trillions of data corpus but pre-train for very limited (i.e., only near-one) epoch, believing in the model’s memory of each individual token is not practical. Even the size of today's large models is far from enough to memorize verbatim. Because of the black-box, we don’t know exactly what the model remembers, however we can try to get clues from some details displayed by the model itself.

\subsection{Surprising Tokens Identification}
\label{subsec:surpring-tokens}
Language models determine what the current token should be based on the preceding text, and \textit{entropy} reveals the model's certainty in deciding the current token should be which one among the entire vocabulary. 

Shannon Entropy\cite{shannon1948mathematical} is widely used as a measurement for the uncertainty within a distribution. Specifically, it is defined as $H(P_X) = - \Sigma_{x \in \mathcal{X}} P_X(x) \log P_X(x)$, where $P_X$ represents a discrete probability distribution across the outcome space $\mathcal{X}$, and its value falls between $0$ and $\log|\mathcal{X}|$ (i.e., occurs when all outcomes are equally like). Generally, lower entropy means the probability distribution is more concentrated, and there's less uncertainty about the outcome; whereas higher entropy indicates a more spread-out distribution with greater uncertainty. Thus in context of language models, Shannon entropy can be used to evaluate the predictability or uncertainty of the text sequences. That is, a \textit{low entropy} distribution in language models implies that there are not so much possible choices, leading to \textit{model's great certain} for the current token. 

Based on these, we define \textit{surprising tokens} as the tokens that 1) model is confident with the prediction of the current token location, which can be reflected from the low entropy value at the index; 2) the model has low probability for predicting this token to be the ground truth one. 
When these two conditions achieve at the same time, the model is confident that the ground truth token is not the current token. 
As a result, upon revealing the ground truth, the model will be surprising. Note that if the model itself is not very certain about what this position should be (i.e., it has a high entropy distribution), the low probability prediction of the ground truth token doesn't signify much, regardless of whether the data has been seen before or not.

\subsection{Adaptive Pre-training Data Detection}
\label{subsec:detection-method}

We introduce an adaptive pre-training data detection method \textbf{\model} that using token probabilities at adaptively selected representative index (aka. \textit{surprising tokens}) of a given input for detection. \model is based on the hypothesis that the seeing \textit{seen} data will \textit{not be that surprising} for the model when compared with seeing \textit{unseen} data. 

\begin{figure}
    \centering
    \subfloat[][All input tokens.]{\includegraphics[width=.48\textwidth]{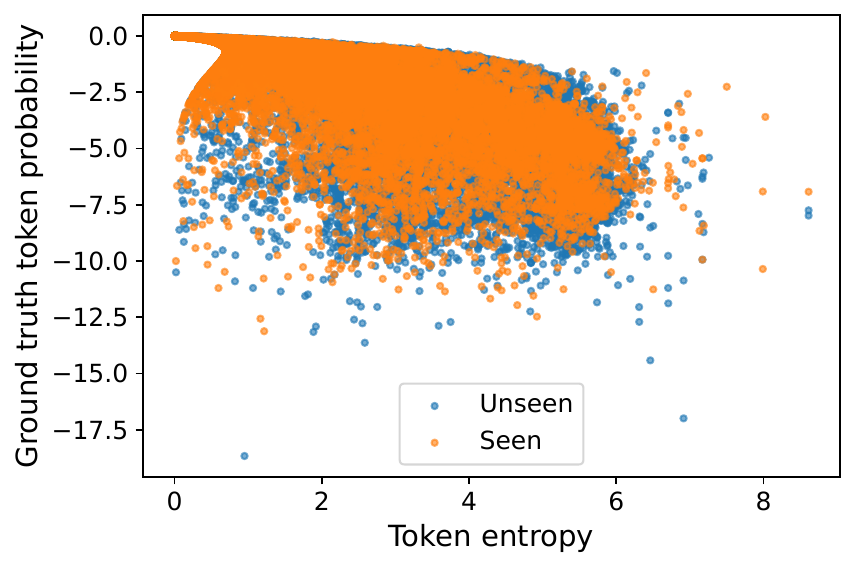}\label{fig:demo1}}
    \hfill
    \subfloat[][tokens have low entropy (<2) and low ground truth log probability (<20-th percentile).]{\includegraphics[width=.48\textwidth]{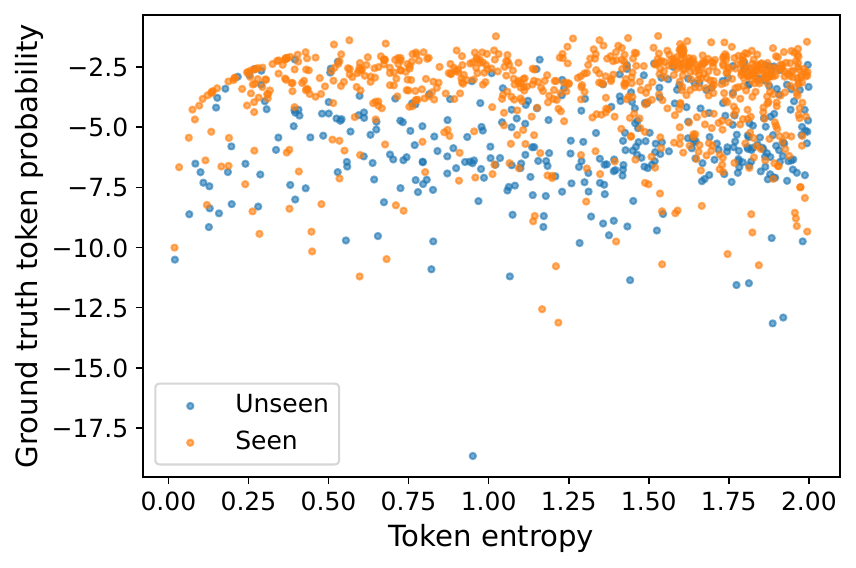}\label{fig:demo2}}
    \caption{The calculated token entropy values and model's prediction log-probability on ground truth tokens for (a) all input tokens (b) tokens have low entropy and low ground truth probability, using GPT-Neo-2.7B on DM Math dataset.}
    \label{fig:demo}
\vspace{-15pt}
\end{figure}
Our inspiration is derived from the observations about the \textit{surprising tokens}: we found that when select representative tokens based on the requirements described in \S\ref{subsec:surpring-tokens}, the average probability of ground truth tokens at the selected indexes of a seen data is higher than it of a unseen data. We empirically demonstrate this phenomenon in Figure~\ref{fig:demo} using an example of GPT-Neo-2.7B model and the DM Math dataset. 
Figure~\ref{fig:demo1} shows the token entropy and ground truth token log probability for all tokens within given inputs, where we are hard to tell the difference between unseen inputs and seen inputs. We notice that in the left bottom part of the figure, which is the range for the \textit{surprising tokens}, tokens from seen inputs and tokens from unseen inputs are more separable. Figure~\ref{fig:demo2} is drawn by limiting the token entropy and ground truth probability(i.e., the left bottom part of Figure~\ref{fig:demo1}). We can clearly see that the tokens from seen inputs have a larger prediction probability compared with tokens from unseen inputs. Therefore, we focus on \textit{surprising tokens} and design \model.

\model is formulated as following. 
Denote a sequence of tokens, which is the input, as $\mathbf{x}=x_1, x_2, x_3, \dots, x_N$. 
Denote a language model as $\mathcal{M}$, which takes some of the token sequence $x_1, x_2, \dots, x_{i-1}$ as input and predict the probability of the next token among the set of vocabulary $\mathcal{V}$ at location $i$. 
We denote $x_i$ as the ground truth token at position $i$. 
The probability distribution of each token position as a $1 \times \mid \mathcal{V}\mid$ vector $P_i$, where $i$ indicates the index in the input sequence $\mathbf{x}$. 
Each entry of the probability vector $P_i$, which represents the model's predicted probability on a specific token candidate, is calculated as $P_{i, v_j} = p(v_j\mid x_{<i}; \mathcal{M}), \forall~v_j\in \mathcal{V}$. 
Denote the ground truth token's log probability as $L_i$, with the probability vector, it can be calculate as $L_i = \log P_{i, x_i}$.
We first obtain the entropy $E_i$ at each index $i$, then select a subset of indexes where entropy is smaller than a certain value $\varepsilon_e$:
\begin{align}
\label{eq:opt-func-stru}
E_i &= -\sum_{v_j\in \mathcal{V}}P_{i,v_j}\log{P_{i,v_j}}, \\
S_e &= \{i \mid  E_i < \varepsilon_e, \forall~i \in \{1, \dots, N\}\}
\end{align}


We further restrict set $S_e$ by choosing indexes with low prediction probability for the ground truth token $x_i$, i.e., $P_{i, x_i}$ is low. We define low probability in terms of $k$-th-percentile of log probabilities $L_1, L_2, \dots, L_N$. The $k$-th percentile is the value $k/100$ of the way from the minimum to the maximum. Denote this value as $L^k$, 
we have
\begin{equation}
    S_p = \{i \mid  L_i < L^k, \forall~i \in \{1, \dots, N\}\}
\end{equation}

Finally, for each given input sequence $\mathbf{x}$ we adaptively locate the set of \textit{surpring tokens} as $S_e\cap S_p$, and calculate the average log probability of the tokens in the joint of these two:
\begin{equation}
    \model (\mathbf{x}, \mathcal{M}) = \frac{1}{\mid S_e\cap S_p\mid} \sum_{i\in S_e\cap S_p} L_i
\end{equation}

To detect whether a piece of text $\mathbf{x}$ is part of the pre-training data, we simply thresholding the value $\model (\mathbf{x}, \mathcal{M})$:
\begin{equation}
    \mathcal{F}(\model(\mathbf{x}, \mathcal{M}); \lambda) = \mathbbm{1}\left[ \model(\mathbf{x}, \mathcal{M}) \geq \lambda \right]
\end{equation}
where $\lambda$ is a case-dependent threshold. We care about the threshold-independent metric in \S \ref{sec:eval} same as previous works. 
The algorithm is summarized in Algorithm \ref{alg:1} in Appendix \ref{app:algo}.

\section{Experiments}
\label{sec:eval}

We evaluate the performance of the proposed \model as well as prior detection baseline methods against large autoregressive models over three benchmarks targeting pre-training data detection. 
We start by describing the datasets, models and baselines used in our experiments.

\subsection{Setups}
\label{subsec:eval-setups}

\paragraph{Datasets.} We perform experiments on three benchmarks, including two existing benchmarks \textbf{WikiMIA}~\cite{shi2024detecting} and \textbf{MIMIR}~\cite{duan2024membership}, as well as a new benchmark \textbf{Dolma-Book} that we developed upon a novel framework. 
WikiMIA~\cite{shi2024detecting} is a dataset that collect event pages from Wikipedia and distinguish training and non-training data based on a specific cutoff date. Moreover, WikiMIA\footnote{WikiMIA from HuggingFace. https://huggingface.co/datasets/swj0419/WikiMIA} provides categorized data based on different sentence lengths (i.e., $32, 64, 128, 256$ words in one input separately), aiming to offer a detailed evaluation of varying input lengths.
MIMIR~\cite{duan2024membership} is benchmark that built based on seven diverse data sources in the Pile dataset~\cite{gao2020pile}. Training and non-training data in this benchmark are directly sampled from the train and test sets of the Pile respectively. Moreover, MIMIR\footnote{MIMIR from HuggingFace. https://huggingface.co/datasets/iamgroot42/mimir} provides data in different source domains with different $n$-gram overlap between the training and non-training data, and we keep all methods to use the $\leq 20\% \ 7$-gram overlap data groups in our evaluation since baseline methods show better performance. 
\paragraph{Dolma-Book Benchmark Construction.} We construct a new benchmark \textbf{Dolma-Book},  which is developed upon a novel LLM framework -- Open Language Model (OLMo)~\cite{OLMo}, to provide further evaluation of our method. 
Since it releases full pre-training data (i.e., Dolma) for the OLMo models, we could easily sample training data from the released data. 
However, we still need to collect non-training data for the detection problem. 
Therefore, to construct our Dolma-Book benchmark: \textbf{Step 1.} Collect non-training book data from Project Gutenberg, which is a repository of over $70$ thousand public domain books that are not protected under U.S. copyright law. Since Dolma collected this archive in April 2023, we set January 1, 2024 as the cutoff date and extract all books after this date directly from the website \href{https://www.gutenberg.org/}{gutenberg.org}\footnote{Use GutenbergPy~\cite{GutenBergPy} to extract books.}. There are 978 books in total. 
\textbf{Step 2.} For the pre-training data, we randomly sample from the groups of book data within Dolma\footnote{Dolma from HuggingFace. https://huggingface.co/datasets/allenai/dolma} for the same number of books with the non-training ones. 
\textbf{Step 3.} The raw books are too long to be inputs for a benchmark, so we need to select parts of them. In order to further compare the detection performance of different text positions within the long-context raw book, we collect and category three datasets for both the training and non-training, called \textit{Dolma-Book-head}, \textit{Dolma-Book-middle} and \textit{Dolma-Book-tail}. We first split each book into segmentations (i.e., after discarding the headers of the book) according to a fixed number of words (i.e., do $1024$ and $512$). Then we take the first segmentation, middle segmentation(i.e., $length //2$) and the last two segmentations into the datasets of \textit{head}, \textit{middle} and \textit{tail} respectively. The procedure is conducted for both training and non-training data creation for further evaluations. 
The data construction pipeline including new data collection and categorization is automatic and is convenient to extend for future time.
\paragraph{Models.} We conduct various experiments on eight large language models among four families. Given that Wikipedia often serves as part of the training corpus for multiple models, we evaluate the performance on WikiMIA against the suite of LLaMA~\cite{touvron2023llama} (7B, 13B), Pythia~\cite{biderman2023pythia} (2.8B, 6.9B, 12B) and GPT-Neo~\cite{gpt-neo} (2.7B) models. For Dolma-Book, we use the two different-size models (1B, 7B) provided by OLMo framework~\cite{OLMo}. Then we use the suite of Pythia and GPT-Neo models for the MIMIR, since these models are trained on the Pile. We also evaluate on the deduped-Pythia model(Table.~\ref{tab:deduped_compare}) to compare the performance. 
\paragraph{Evaluation Metrics.} Consistent with previous works, we primarily report the AUC-ROC score (the area under ROC curve) for performance evaluation, and show the TPR@low\% FPR scores. 
\paragraph{Baselines.} We consider six baselines including the standard reference-free and reference-based methods to compare the performance. 
The LOSS Attack method (\textbf{PPL}) \cite{yeom2018privacy} uses the model's computed loss of an input as the perplexity to predict the membership. 
Neighbor method (\textbf{Neighbor}) \cite{mattern-etal-2023-membership} leverages an estimation of the loss curvature to detect membership. 
Min-K\% Prob method (\textbf{MinK}) \cite{shi2024detecting} selects $k\%$ tokens which have the lowest likelihood to compute the score for detection.
Reference-based method (\textbf{Ref}) proposed in \cite{carlini2021extracting} trains a small reference language model on the same data and uses the perplexity obtained from the reference model for calibration. 
In evaluation, we use the smaller version of each family as their reference model, i.e., LLaMA-160M for LLaMA models, Pythia-70M for Pythia models, GPT-Neo-125M for GPT-Neo-2.8B and OLMo-1B for OLMo-7B. 
Similar to reference-based method, \textbf{Zlib} method~\cite{carlini2021extracting} uses zlib compression entropy as reference, and \textbf{Lowercase}~\cite{carlini2021extracting} uses lowercased example perplexity as the reference.


\subsection{Evaluation Results}
\label{subsec:eval_main}


\begin{table*}[t]
\caption{AUC-ROC scores for pre-training data detection on WikiMIA and various models across \model and baseline methods. Show results of the input length $32$ and $128$ respectively, with highlighting best results as bold. }\label{tab:wikimia_results}
\begin{center} 
\setlength{\tabcolsep}{0.8pt}
\begin{tabularx}{\textwidth}{l *{14}{>{\centering\arraybackslash}X}@{}}
    \toprule
     \multirow{2}{*}{} & \multicolumn{2}{c}{LLaMA-7B} & \multicolumn{2}{c}{LLaMA-13B} & \multicolumn{2}{c}{Pythia-2.8B} & \multicolumn{2}{c}{Pythia-6.9B} & \multicolumn{2}{c}{Pythia-12B} & \multicolumn{2}{c}{GPT-neo-2.7B} \\
    \cmidrule(lr){2-3}  \cmidrule(lr){4-5} \cmidrule(lr){6-7} \cmidrule(lr){8-9} \cmidrule(lr){10-11} \cmidrule(lr){12-13}
    \textbf{Method} & 32 & 128
    & 32 & 128
    & 32 & 128
    & 32 & 128
    & 32 & 128
    & 32 & 128
    \\
    \midrule
        PPL & 0.661 & 0.666 & 0.675 & 0.678 & 0.614 & 0.628 & 0.638 & 0.651 & 0.654 & 0.658 & 0.620 & 0.640 \\
        Ref & 0.595 & 0.581 & 0.608 & 0.594 & 0.613 & 0.596 & 0.636 & 0.633 & 0.651 & 0.639 & 0.605 & 0.594 \\
        Lower & 0.604 & 0.591 & 0.640 & 0.606 & 0.609 & 0.595 & 0.622 & 0.605 & 0.647 & 0.614 & 0.611 & 0.614 \\
        Zlib & 0.667 & 0.683 & 0.678 & 0.697 & 0.622 & 0.650 & 0.643 & 0.676 & 0.658 & 0.678 & 0.625 & 0.662 \\
        Neighbor & 0.629 & 0.616 & 0.642 & 0.652 & 0.599 & 0.619 & 0.625 & 0.634 & 0.630 & 0.634 & 0.605 & 0.629 \\
        MinK & 0.651 & 0.697 & 0.668 & 0.715 & 0.617 & 0.668 & 0.663 & 0.695 & 0.681 & 0.707 & 0.639 & 0.683 \\
        \model & \textbf{0.864} & \textbf{0.832} & \textbf{0.868} & \textbf{0.830} & \textbf{0.633} & \textbf{0.671} & \textbf{0.681} & \textbf{0.698} & \textbf{0.698} & \textbf{0.710} & \textbf{0.662} & \textbf{0.694} \\
    \bottomrule
\end{tabularx}
\end{center}
\end{table*}

\begin{table*}[t]
\begin{minipage}{\textwidth}
\caption{AUC-ROC results for pre-training data detection on three datasets (\textit{head}, \textit{middle}, \textit{tail}) in Dolma-Book, using different sizes of OLMo models across \model and baseline methods. Each input text has the word length to be $1024$. Highlight best results as bold.}\label{tab:dolma_results}
\centering
\begin{tabularx}{\textwidth}{l *{8}{>{\centering\arraybackslash}X}@{}}
    \toprule
     \multirow{2}{*}{} & \multicolumn{2}{c}{Dolma-Book-head} & \multicolumn{2}{c}{Dolma-Book-middle} & \multicolumn{2}{c}{Dolma-Book-tail}  \\
    \cmidrule(lr){2-3}  \cmidrule(lr){4-5} \cmidrule(lr){6-7}
    \textbf{Method\footnote{We skip neighbor method due to out-of-memory.}} & OLMo-1B & OLMo-7B
    & OLMo-1B & OLMo-7B
    & OLMo-1B & OLMo-7B
    \\
    \midrule
        PPL & 0.609 & 0.596 & 0.618 & 0.616 & 0.575 & 0.576 \\
        Ref\footnote{OLMo-1B is the smallest model in OLMo framework, so we skip reference method for OLMo-1B.} & - & 0.337 & - & 0.434 & - & 0.447 \\
        Lower & 0.471 & 0.469 & 0.504 & 0.497 & 0.473 & 0.469 \\
        Zlib & \textbf{0.741} & \textbf{0.739} & 0.472 & 0.475 & 0.526 & 0.527 \\
        MinK & 0.579 & 0.586 & 0.583 & 0.589 & 0.564 & 0.571 \\
        \model & 0.634 & 0.627 & \textbf{0.640} & \textbf{0.649} & \textbf{0.602} & \textbf{0.608} \\
    \bottomrule
\end{tabularx}
\end{minipage}
\end{table*}

Table.~\ref{tab:wikimia_results},~\ref{tab:dolma_results},~\ref{tab:mimir_results} show the results of our method across three benchmarks on various models and the comparisons with the baseline methods. For all experiments in evaluation, we apply grid search for the best hyperparameters, i.e., $\varepsilon_e$ is from $0.5$ to $10$ with step size $0.5$, and $k$ is from $10$ to $100$ with step size $10$. Same with the baseline MinK, the hyperparameter can be determined by a held-out validation set. We show more analysis about our hyperparemeters in \S \ref{subsec:eval-ablation}. For each of the baseline, we basically apply setups according to their papers and implementations. And for MinK, we report the results under the best parameter value for a fair comparison. 

The AUC-ROC results for \textbf{WikiMIA} benchmark is shown in Table.~\ref{tab:wikimia_results}. In addition to different kinds of LLMs, we also evaluate the detection performance on various length of the input. Results of $32$ and $128$ words length are displayed here, and the results of the input length to be $64$ and $256$ are in Appendix.~\ref{app:wikimia}. 
As shown in the table, our method can achieve a huge enhancement with maximum of $29.5\%$, and consistently outperform baselines across different input lengths (average improvement of $11.4\%$ for length-32 and average $6.4\%$ for length-128).
In general, the performance increases as the input text length increases, except for LLaMA models which display a dramatic advantage in shorter inputs. Also, according to the results from the suite of Pythia models, the performance will get improved with larger size of models regardless of the input length. 

Table.~\ref{tab:dolma_results} shows the AUC-ROC results for our \textbf{Dolma-Book} benchmark. Given the considerable length of the raw book, we conduct separate comparisons of input data from the \textit{head}, \textit{middle} and \textit{tail} segmentations(details in \S \ref{subsec:eval-setups}) of the book to show a better evaluation of detection performance across different text positions within the long-context book data. As shown in Table.~\ref{tab:dolma_results}, our method outperforms other baseline methods on both \textit{Dolma-Book-middle} and \textit{Dolma-Book-tail} with an improvement of $5.2\%$ and $5.6\%$ from the best baseline AUC scores respectively. 
For \textit{Dolma-Book-head}, Our method performs the best except for Zlib. The \textit{head} segmentation of a book could contains a certain degree of garbled characters (including formatting, list of names, table of contents, etc.), which is likely to make Zlib compression calibration effective.
Moreover, basically for all methods, the \textit{middle} of book as input can obtain best detection performance, and \textit{tail} gets the worst. 
Appendix.~\ref{app:dolma} show more results about $512$ word-length of inputs. 

Table.~\ref{tab:mimir_results} shows the results of using \textbf{MIMIR} benchmark. Among seven different datasets, our method basically outperforms other baselines or reach a comparable results (second best beyond Zlib for Github). For some datasets like DM Math, our method achieves an enhancement from the best previous method's (Lowercase) $92.8\%$ to $94.2\%$, but we may also obtain modest improvement on some datasets such as HackerNews.
\paragraph{TPR@lower FPR.} 
Following prior works, we additionally report the detection performance in terms of the true positive rates (TPR) under low false positive rates (FPR). Figure.~\ref{fig:tpr_fpr} shows the ROC curve of WiKIMIA data and LLaMA-13B model across various methods. Our method significantly improve existing approaches through varying threshold, e.g., we improve TPR from $0.275$ to $0.431$ at $5\%$ FPR, and from $0.196$ to $0.412$ at $1\%$ FPR. Table.~\ref{tab:tpr_fpr1_wikimia} - \ref{tab:tpr_fpr5_mimir} in Appendix.~\ref{app:tpr_fpr} display the values of TPR@low\% FPRs (i.e., $1\%$, $5\%$ and $10\%$) for different benchmarks and LLMs.

\begin{table*}[t]
\caption{AUC-ROC results for pre-training data detection on seven datasets in MIMIR, using various models across SURP and baseline methods. Highlight best results as bold.}\label{tab:mimir_results}
\begin{center} 
\scriptsize
\setlength{\tabcolsep}{0.7pt}
\begin{tabularx}{\textwidth}{l *{18}{>{\centering\arraybackslash}X}@{}}
    \toprule
    \multirow{2}{*}{}  & \multicolumn{4}{c}{Wikipedia} & \multicolumn{4}{c}{Github} & \multicolumn{4}{c}{Pile CC} & \multicolumn{4}{c}{PubMed Central} \\
    \cmidrule(lr){2-5}  \cmidrule(lr){6-9} \cmidrule(lr){10-13} \cmidrule(lr){14-17}
    \textbf{Method} & Pythia-2.8B & Pythia-6.9B & Pythia-12B & Neo-2.7B
    & Pythia-2.8B & Pythia-6.9B & Pythia-12B & Neo-2.7B
    & Pythia-2.8B & Pythia-6.9B & Pythia-12B & Neo-2.7B
    & Pythia-2.8B & Pythia-6.9B & Pythia-12B & Neo-2.7B
    \\
    \midrule
        PPL & 0.663 & 0.679 & 0.690 & 0.657 & 0.880 & 0.889 & 0.895 & 0.873 & 0.549 & 0.561 & 0.565 & 0.553 & 0.781 & 0.782 & 0.781 & 0.798 \\
        Ref & 0.561 & 0.582 & 0.593 & 0.546 & 0.418 & 0.426 & 0.430 & 0.416 & 0.530 & 0.540 & 0.545 & 0.544 & 0.392 & 0.396 & 0.397 & 0.351 \\
        Lower & 0.648 & 0.669 & 0.676 & 0.660 & 0.869 & 0.881 & 0.889 & 0.881 & 0.548 & 0.561 & \textbf{0.572} & 0.545 & 0.724 & 0.728 & 0.725 & 0.734 \\
        Zlib & 0.630 & 0.649 & 0.661 & 0.623 & \textbf{0.907} & \textbf{0.914} & \textbf{0.917} & \textbf{0.899} & 0.537 & 0.548 & 0.551 & 0.540 & 0.771 & 0.773 & 0.772 & 0.786 \\
        Neighbor & 0.572 & 0.567 & 0.607 & 0.567 & 0.868 & 0.872 & 0.881 & 0.866 & 0.480 & 0.491 & 0.482 & 0.483 & 0.617 & 0.624 & 0.611 & 0.625 \\
        MinK & 0.657 & 0.679 & 0.695 & 0.649 & 0.879 & 0.889 & 0.895 & 0.872 & 0.547 & 0.562 & 0.563 & 0.549 & 0.781 & 0.786 & 0.787 & 0.796 \\
        \model & \textbf{0.669} & \textbf{0.688} & \textbf{0.702} & \textbf{0.660} & 0.881 & 0.890 & 0.896 & 0.874 & \textbf{0.553} & \textbf{0.564} & 0.571 & \textbf{0.556} & \textbf{0.789} & \textbf{0.793} & \textbf{0.793} & \textbf{0.805} \\
        \vspace{-.6em} \\
    \toprule
    \multirow{2}{*}{}  & & \multicolumn{4}{c}{ArXiv} & & \multicolumn{4}{c}{DM Math} & & \multicolumn{4}{c}{HackerNews} &  \\
    \cmidrule(lr){3-6}  \cmidrule(lr){8-11} \cmidrule(lr){13-16}
    \textbf{Method} &  & Pythia-2.8B & Pythia-6.9B & Pythia-12B & Neo-2.7B
     & & Pythia-2.8B & Pythia-6.9B & Pythia-12B & Neo-2.7B
     & & Pythia-2.8B & Pythia-6.9B & Pythia-12B & Neo-2.7B & 
    \\
    \midrule
        PPL         &  & 0.780 & 0.791 & 0.795 & 0.790 &  & 0.919 & 0.920 & 0.919 & 0.930 &  & 0.606 & 0.613 & 0.621 & 0.592 \\
        Ref         &  & 0.565 & 0.586 & 0.596 & 0.548 &  & 0.377 & 0.365 & 0.363 & 0.419 &  & 0.522 & 0.532 & 0.550 & 0.517 \\
        Lower       &  & 0.737 & 0.746 & 0.753 & 0.736 &  & 0.928 & 0.928 & 0.902 & 0.839 &  & 0.540 & 0.541 & 0.547 & 0.539 \\
        Zlib        &  & 0.775 & 0.784 & 0.787 & 0.784 &  & 0.811 & 0.813 & 0.812 & 0.812 &  & 0.594 & 0.598 & 0.604 & 0.587 \\
        Neighbor    &  & 0.664 & 0.670 & 0.662 & 0.653 &  & 0.752 & 0.775 & 0.770 & 0.763 &  & 0.538 & 0.531 & 0.570 & 0.545 \\
        MinK        &  & 0.752 & 0.767 & 0.778 & 0.760 &  & 0.926 & 0.925 & 0.924 & 0.933 &  & 0.582 & 0.592 & 0.605 & 0.572 \\
        \model      &  & \textbf{0.784} & \textbf{0.793} & \textbf{0.798} & \textbf{0.791} &  & \textbf{0.942} & \textbf{0.942} & \textbf{0.936} & \textbf{0.945} &  & \textbf{0.606} & \textbf{0.614} & \textbf{0.624} & \textbf{0.594} \\
    \bottomrule
\end{tabularx}
\end{center}
\vskip -0.1in
\end{table*}

\subsection{Ablation Studies}
\label{subsec:eval-ablation}


\begin{figure*}[]
\centering
    \begin{minipage}[t]{0.40\linewidth}
    \centering
    \includegraphics[width=1\columnwidth]{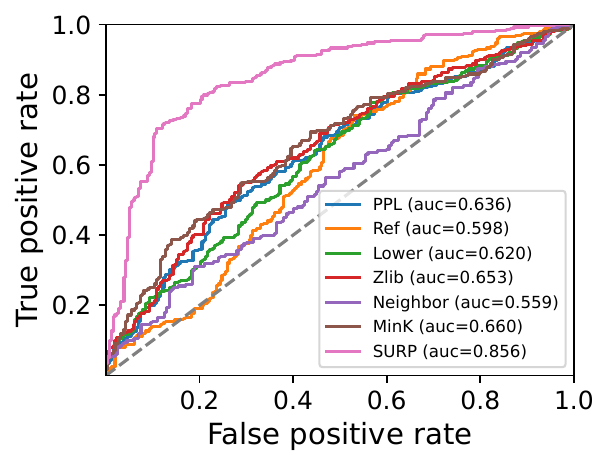}
    \captionof{figure}{The ROC curve for LLaMA-13B on WikiMIA length-$64$ dataset.} 
     \label{fig:tpr_fpr}
    \end{minipage}
    \hfill
    \begin{minipage}[t]{0.55\textwidth}
    \centering
    \includegraphics[width=1\columnwidth]{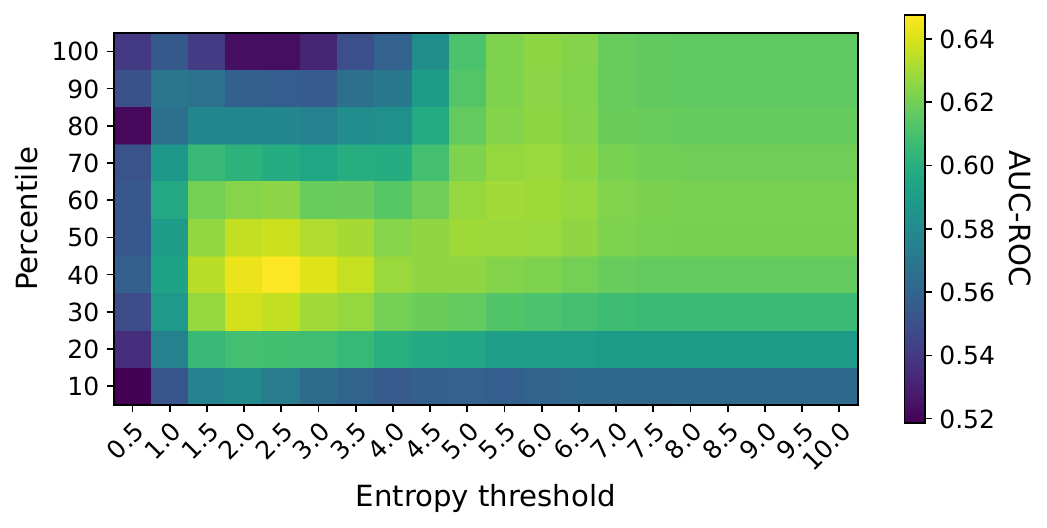}
    \caption{Heatmap to show the AUC scores of different hyperparameters, using OLMo-7B on Dolma-Book-middle dataset.}
    \label{fig:param}
    \end{minipage}
\end{figure*}

\paragraph{Analysis about the hyperparameters.} 

We have two parameters, one is the entropy threshold $\varepsilon_e$, one is the low probability threshold $k$-percentile. For giving more analysis, we show the heatmap of AUC scores in Figure \ref{fig:param}. 
Figure \ref{fig:param} shows the AUC score when using different entropy threshold and percentile pairs, with OLMo-7B model and Dolma-Book-middle dataset. We can find that the best AUC score happens when the $\varepsilon_e = 2.5$ and the $k=40$. We can see clearly that both entropy threshold and low ground truth probability are important: if only use low entropy threshold $\varepsilon_e$ (the top row in figure), the AUC score could as low as 0.52 (with $\varepsilon_e = 2.5$ and $k=100$); if only use small $k$ (the right most column), the AUC score is around 0.56 (with $\varepsilon_e = 10.0$ and $k=10$). 
Similar phenomena is shown from Figure~\ref{fig:other_heatmap} in Appendix~\ref{app:heatmap} using Dolma-Book-tail dataset, which achieves best AUC when entropy threshold is low ($\varepsilon_e=2.0$) and $k=30$.

\paragraph{Deduped model cannot decrease the advantage. }
To see the influence of duplicates in training data, we compare the performance of model trained on regular dataset and deduplicate dataset. We evaluate the performance of our method and baselines using Pythia-2.8B and deduped-Pythia-2.8B, see Table.~\ref{tab:deduped_compare} in Appendix~\ref{app:deduped}. For using the deduped model, our method also achieve the best results comparing with other baselines over various datasets (except for Github as second best after Zlib). Also, our method achieve similar performance on the original model and deduplicate model (i.e., non-deduped slightly higher in Wikipedia, ArXiv and DM Math). 

\paragraph{Influence of data distribution shifts. }
Data distribution shifts could occur in current dataset for pre-training detection, as in many datasets, the unseen data consists of newly added data after the trained model was released. This brings a potential data distribution shifts. To see the influence of data distribution shifts, we evaluate the performance on temporal datasets of Wikipedia and ArXiv using the datasets provided by \cite{duan2024membership}. The temporal datasets contain data distribution shifts between training data and non-training data, as they are created with texts in different time periods. When comparing the results between with-distribution-shifts (i.e., temporal Wikipedia/ArXiv) and without-distribution-shifts (i.e., original Wikipedia/ArXiv), as shown in Table \ref{tab:data_distribution_results} in Appendix~\ref{app:temporal}, we see no consistent performance improvement on the temporal datasets compared with common ones, as Wikipedia-temporal improves a lot from $68.8\%$ to $75.6\%$ while ArXiv-temporal performs a negative impact from $79.3\%$ to $73.4\%$. 



\section{Conclusion}
\label{sec:conclusion}
We provide an adaptive pre-training data detection method, which alleviate the reliance of verbatim memorization and effectively amplify the identification. 
Our method adaptively identify surprising tokens of the input, and detect whether the input is seen or not based on the hypothesis that seen data is less surprising for the model compared with unseen data. We construct data benchmark Dolma-Book, and evaluate our method on our benchmark and other existing benchmarks.
Our work further suggests the impractical of simple application of traditional MIAs for pre-training data detection in LLMs, and we encourage the development of new strategies considering completly outside the verbatim memorization for achieving effective and robust pre-training data detection. We expect to see the positive social impact of our method and other work in line, especially in protecting private and copyright data. 
\newpage
\bibliography{reference} 






\end{document}